\documentclass{article} 
\usepackage{microtype}
\usepackage{graphicx}
\usepackage{hyperref}
\usepackage[accepted]{icml2018}

\usepackage{url}
\usepackage{natbib} 
\bibpunct{(}{)}{,}{a}{}{;}
\usepackage{amsmath} 
\usepackage{amssymb} 
\usepackage{mathrsfs}
\usepackage{pgfplots}
\usepackage{xcolor}
\usepackage{pgfplots}
\usepgfplotslibrary{groupplots}
\usetikzlibrary{pgfplots.groupplots}
\newlength\figH
\newlength\figW
\newlength\textSize

\newcommand{\weight}{w}
\newcommand{\weights}{\boldsymbol{\weight}}

\newcommand{\initweights}{\weights^0}

\newcommand{\funcnet}[1][k]{f_{#1}}

\newcommand{\group}[1][g]{{\mathcal G}_{#1}}

\newcommand{\source}{\mathcal{S}}
\newcommand{\sourceweight}{\weight}
\newcommand{\sourceweights}{\weights_{\source}}
\newcommand{\newweights}{\weights_{\bar{\source}}}
\newcommand{\sourceinitweights}{\weights^{0}_{\source}}
\newcommand{\sourceinitweight}{\weight^{0}}

\newcommand{\norm}[2][2]{\left\|#2\right\|_{#1}}

\newcommand{\Rset}{\mathbb{R}}

\newcommand{\eqdef}{\stackrel{\triangle}{=}}

\makeatletter
\def\xinput{%
   \@ifnextchar[%
     {\xinput@i}
     {\xinput@i[j]}%
}
\def\xinput@i[#1]{%
   \@ifnextchar[%
     {\xinput@ii{#1}}
     {\xinput@ii{#1}[i]}%
}
\def\xinput@ii#1[#2]{%
  \@ifempty{#2}%
    {x_{#1}^{\phantom{()}}}
    {\@ifempty{#1}%
    {x_{j}^{(#2)}}
    {x_{#1}^{(#2)}}}%
}
\def\xinputs{%
   \@ifnextchar[%
     {\xinputs@i}
     {\xinputs@i[i]}%
}
\def\xinputs@i[#1]{%
  \@ifempty{#1}%
    {\boldsymbol{x}}
    {\boldsymbol{x}^{(#1)}}%
}
\def\youtput{%
   \@ifnextchar[%
     {\youtput@i}
     {\youtput@i[k]}%
}
\def\youtput@i[#1]{%
   \@ifnextchar[%
     {\youtput@ii{#1}}
     {\youtput@ii{#1}[i]}%
}
\def\youtput@ii#1[#2]{%
  \@ifempty{#2}%
    {y_{#1}^{\phantom{()}}}
    {\@ifempty{#1}%
    {y_{j}^{(#2)}}
    {y_{#1}^{(#2)}}}%
}
\def\youtputs{%
   \@ifnextchar[%
     {\youtputs@i}
     {\youtputs@i[i]}%
}
\def\youtputs@i[#1]{%
  \@ifempty{#1}%
    {\boldsymbol{y}}
    {\boldsymbol{y}^{(#1)}}%
}
\makeatother

\usepackage{ulem}
\icmltitlerunning{Explicit Inductive Bias for Transfer Learning with Convolutional Networks}

\begin{document}

\twocolumn[
\icmltitle{Explicit Inductive Bias for Transfer Learning with Convolutional Networks}
        
\begin{icmlauthorlist}
\icmlauthor{Xuhong LI}{to}
\icmlauthor{Yves GRANDVALET}{to}
\icmlauthor{Franck DAVOINE}{to}
\icmlaffiliation{to}{Sorbonne universit\'es, Universit\'e de technologie de Compi\`egne, CNRS, Heudiasyc, UMR 7253, Compi\`egne, France}
\end{icmlauthorlist}

\icmlcorrespondingauthor{Xuhong LI}{xuhong.li@hds.utc.fr}

\icmlkeywords{Machine Learning, transfer learning, regularization}

\vskip 0.3in
]

\printAffiliationsAndNotice{} 

\begin{abstract}
In inductive transfer learning, fine-tuning pre-trained convolutional networks substantially outperforms training from scratch.
When using fine-tuning, the underlying assumption is that the pre-trained model extracts generic features, which are at least partially relevant for solving the target task, but would be difficult to extract from the limited amount of data available on the target task.
However, besides the initialization with the pre-trained model and the early stopping, there is no mechanism in fine-tuning for retaining the features learned on the source task.
In this paper, we investigate several regularization schemes that explicitly promote the similarity of the final solution with the initial model.
We show the benefit of having an explicit inductive bias towards the initial model, and we eventually recommend a simple $L^2$ penalty with the pre-trained model being a reference as the baseline of penalty for transfer learning tasks.
\end{abstract}

\section{Introduction}
\label{section_in}

It is now well known that modern convolutional neural networks \citep[e.g.][]{krizhevsky2012imagenet, simonyan2015very, he2016deep, szegedy2016rethinking} can achieve remarkable performance on large-scale image databases, e.g. ImageNet \citep{deng2009imagenet} and Places 365 \citep{zhou2017places},
but it is really dissatisfying to see the vast amounts of data, computing time and power consumption that are necessary to train deep networks.
Fortunately, such convolutional networks, once trained on a large database, can be refined to solve related but different visual tasks by means of transfer learning, using fine-tuning \citep{yosinski2014transferable, simonyan2015very}.

Some form of knowledge is believed to be extracted by learning from the large-scale database of the source task and this knowledge is then transferred to the target task by initializing the network with the pre-trained parameters.
However, we will show in the experimental section that some parameters may be driven far away from their initial values during fine-tuning.
This leads to important losses of the initial knowledge that is assumed to be relevant for the targeted problem.

In order to help preserve the knowledge embedded in the initial network, we consider a series of other parameter regularization methods during fine-tuning.
We argue that the standard $L^2$ regularization, which drives the parameters towards the origin, is not adequate in the framework of transfer learning, where the initial values provide a more sensible reference point than the origin.
This simple modification keeps the original control of overfitting, by constraining the effective search space around the initial solution, while encouraging committing to the acquired knowledge.
We show that it has noticeable effects in inductive transfer learning scenarios.

This paper copes with the inconsistency that still prevails in transfer learning scenarios, where the model is initialized with some parameters, while the abuse of $L^2$ regularization encourages departing from these initial values.
We thus advocate for a coherent parameter regularization approach, where the pre-trained model is both used as the starting point of the optimization process and as the reference in the penalty that encodes an explicit inductive bias.
This type of penalty will be designated with \textit{SP} to recall that they encourage similarity with the {\em starting point} of the fine-tuning process.
We evaluate regularizers based on the $L^2$, Lasso and Group-Lasso penalties, which can freeze some individual parameters, or groups of parameters, to the pre-trained parameters.
Fisher information is also taken into account when we test $L^2$\textit{-SP} and Group-Lasso\textit{-SP} approaches.
Our experiments indicate that all tested parameter regularization methods using the pre-trained parameters as a reference get an edge over the standard $L^2$ weight decay approach.
We eventually recommend using $L^2$\textit{-SP} as the standard baseline for solving transfer learning tasks and benchmarking new algorithms.


\section{Related Work}
\label{section:rw}

In this section, we recall the approaches to inductive transfer learning in convolutional networks. 
We focus on approaches that also encourage similarity (of features or parameters) on different models.
Our proposal departs either by the goal pursued or by the type of model used.

\subsection{Shrinking Toward Chosen Parameters}
Regularization has been a means to build shrinkage estimators for decades. 
Shrinking towards zero is the most common form of shrinkage, but shrinking towards adaptively chosen targets has been around for some time, starting with Stein shrinkage \citep[see e.g.][chapter~5]{lehmann1998theory}, where it can be related to empirical Bayes arguments. In transfer learning, it has been used in maximum entropy models \citep{chelba2006adaptation} or SVM \citep{yang2007adapting,aytar2011tabula, tommasi2014learning}.
These approaches were shown to outperform standard $L^2$ regularization with limited labeled data in the target task \citep{aytar2011tabula, tommasi2014learning}.

These relatives differ from the application to deep networks in several respects, the more important one being that
they consider a fixed representation, where transfer learning aims at producing similar classification parameters in that space, that is, similar classification rules.
For deep networks, transfer usually aims at learning similar representations upon which classification parameters will be learned from scratch. 
Hence, even though the techniques we discuss here are very similar regarding the analytical form of the regularizers, they operate on parameters having a very different role.

\subsection{Transfer Learning for Deep Networks}

Regarding transfer learning, we follow here the nomenclature of \citet{pan2010survey}, who categorized several types of transfer learning according to domain
and task settings during the transfer.
A domain corresponds to the feature space and its distribution, whereas a task corresponds to the label space and its conditional distribution with respect to features.
The initial learning problem is defined on the source domain and source task, whereas the new learning problem is defined on the target domain and the target task.

In the typology of \citeauthor{pan2010survey}, we consider the inductive transfer learning setting, where the target domain is identical to the source domain, and the target task is different from the source task.
We furthermore focus on the case where a vast amount of data was available for training on the source problem, and some limited amount of labeled data is available for solving the target problem.
Under this setting, we aim at improving the performance on the target problem through parameter regularization methods that explicitly encourage the similarity of the solutions to the target and source problems.
Note that, though we refer here to problems that were formalized or popularized after \citep{pan2010survey}, such as lifelong learning, \citeauthor{pan2010survey}'s typology remains valid.


\subsubsection{Representation Transfer}
\citet{donahue2014decaf} repurposed features extracted from different layers of the pre-trained AlexNet of \citet{krizhevsky2012imagenet} and plugged them into an SVM or a logistic regression classifier.
This approach outperformed the state of the art of that time on the Caltech-101 database \citep{fei2006one}.
Later, \citet{yosinski2014transferable} showed that fine-tuning the whole AlexNet resulted in better performance than using the network as a static feature extractor.
Fine-tuning pre-trained VGG \citep{simonyan2015very} on the image classification task of VOC-2012 \citep{everingham2010pascal} and Caltech 256 \citep{griffin2007caltech} achieved the best results of that time.

\citet{ge2017borrowing} proposed a scheme for selecting a subset of images from the source problem that have similar local features to those in the target problem and then jointly fine-tuned a pre-trained convolutional network.
Besides image classification, many procedures for object detection \citep{girshick2014rich, redmon2016you, ren2015faster} and image segmentation \citep{long2015fully, chen2017deeplab, zhao2017pyramid} have been proposed relying on fine-tuning to improve over training from scratch.
These approaches showed promising results in a challenging transfer learning setup, as going from classification to object detection or image segmentation requires rather heavy modifications of the architecture of the network.

The success of transfer learning with convolutional networks relies on the generality of the learned representations that have been constructed from a large database like ImageNet.
\citet{yosinski2014transferable} also quantified the transferability of these pieces of information in different layers, e.g. the first layers learn general features, the middle layers learn high-level semantic features and the last layers learn the features that are very specific to a particular task.
That can be also noticed by the visualization of features \citep{zeiler2014visualizing}. 
Overall, the learned representations can be conveyed to related but different domains and the parameters in the network are reusable for different tasks.

\subsubsection{Regularizers in Related Learning Setups}

In lifelong learning \citep{thrun1995lifelong, pentina2015lifelong}, where a series of tasks is learned sequentially by a single model, the knowledge extracted from the previous tasks may be lost as new tasks are learned, resulting in what is known as catastrophic forgetting.
In order to achieve a good performance on all tasks, \citet{li2017learning} proposed to 
use the outputs of the target examples, computed by the original network on the source task, to define a learning scheme preserving the memory of the source tasks when training on the target task.
They also tried to preserve the pre-trained parameters instead of the outputs of examples but they did not obtain interesting results.

\citet{kirkpatrick2017overcoming} developed a similar approach with success.
They get sensible improvements by measuring the sensitivity of the parameters of the network learned on the source data thanks to the Fisher information.
The Fisher information matrix defines a metric in parameter space that is used in their regularizer  to preserve the representation learned on the source data, thereby retaining the knowledge acquired on the previous tasks. 
This scheme, named elastic weight consolidation, was shown to avoid forgetting, but fine-tuning with plain stochastic gradient descent was more effective than elastic weight consolidation for learning new tasks.
Hence, elastic weight consolidation may be thought as being inadequate for transfer learning, where performance is only measured on the target task. 
We will show that this conclusion is not appropriate in typical transfer learning scenarios with few target examples.


In domain adaptation \citep{long2015learning}, where the target domain differs from the source domain whereas the target task is identical to the source task and no (or few) target examples are labeled, most approaches are searching for a common representation space for source and target domains to reduce domain shift.
\citet{rozantsev2016beyond} proposed a parameter regularization scheme for encouraging the similarity of the representations of the source and the target domains. 
Their regularizer encourages similar source and target parameters, up to a linear transformation.
Still in domain adaptation, besides vision, encouraging similar parameters in deep networks has been proposed in speaker adaptation problems \citep{liao2013speaker, ochiai2014speaker} and neural machine translation \citep{barone2017regularization}, where it proved to be helpful.

The $L^2$\textit{-SP} regularizer was used independently by \citet{grachten2017strategies}  for transfer in vision application, but where they used a random reinitialization of parameters.
For convex optimization problems, this is equivalent to fine-tuning with $L^2$\textit{-SP}, but we are obviously not in that situation.
\citet{grachten2017strategies} conclude that their strategy behaves similarly to learning from scratch. 
We will show that using the starting point as an initialization of the fine-tuning process {\em and} as the reference in the regularizer improves results consistently upon the standard fine-tuning process.


\section{Regularizers for Fine-Tuning}
\label{section:ta}

In this section, we detail the penalties we consider for fine-tuning.
Parameter regularization is critical when learning from small databases.
When learning from scratch, regularization is aimed at facilitating optimization and avoiding overfitting, by implicitly restricting the capacity of the network, that is, the effective size of the search space.
In transfer learning, the role of regularization is similar, but the starting point of the fine-tuning process conveys information that pertains to the source problem (domain and task).
Hence, the network capacity has not to be restricted blindly: the pre-trained model sets a reference that can be used to define the functional space effectively explored during fine-tuning.

Since we are using early stopping, fine-tuning a pre-trained model is an implicit form of inductive bias towards the initial solution.
We explore here how a coherent explicit inductive bias, encoded by a regularization term, affects the training process.
Section \ref{sec:experiments} shows that all such schemes get an edge over the standard approaches that either use weight decay or freeze part of the network for preserving the low-level representations that are built in the first layers of the network.

Let $\weights \in \Rset^{n}$ be the parameter vector containing all the network parameters that are to be adapted to the target task.
The regularized objective function $\tilde{J}$ that is to be optimized is the 
sum of the standard objective function $J$ and the regularizer $\Omega(\weights)$.
In our experiments, $J$ is the negative log-likelihood, so that the 
criterion $\tilde{J}$ could be interpreted in terms of maximum {\em a 
posteriori} estimation, where the 
regularizer $\Omega(\weights)$ would act 
as the log prior of $\weights$.
More generally, the minimizer of $\tilde{J}$ is a trade-off between the data-fitting term and the regularization term.

\paragraph{\boldmath${L^2}$ penalty}
The current baseline penalty for transfer learning is the usual $L^2$ penalty, also known as weight decay, since it drives the weights of the network to zero:
\begin{equation}\label{eq:L2}
  \Omega(\weights) = \frac{\alpha}{2} \norm{\weights}^2
  \enspace,
\end{equation}
where  $\alpha$ is the regularization parameter setting the strength of the penalty and $\norm[p]{\cdot}$ is the $p$-norm of a vector. 

\paragraph{\boldmath$L^2$\textit{-SP}}
Let $\initweights$ be the parameter vector of the model pre-trained on the source problem, acting as the starting point (\textit{-SP}) in fine-tuning.
Using this initial vector as the reference in the $L^2$ penalty, we get:
\begin{equation}\label{eq:L2-SP}
  \Omega(\weights) = \frac{\alpha}{2} \norm{\weights-\initweights}^2
  \enspace.
\end{equation}
Typically, the transfer to a target task requires some modifications of the network architecture used for the source task, such as on the last layer used for predicting the outputs.
Then, there is no one-to-one mapping between $\weights$ and $\initweights$, and we use two penalties: one for the part of the target network that shares the architecture of the source network, denoted $\sourceweights$, the other one for the novel part, denoted  $\newweights$. 
The compound penalty then becomes:
\begin{equation}\label{eq:L2-SP-full}
  \Omega(\weights) = \frac{\alpha}{2} \norm{\sourceweights-\sourceinitweights}^2 + \frac{\beta}{2}  \norm{\newweights}^2
  \enspace.
\end{equation}
%

\paragraph{\boldmath$L^2$\textit{-SP-Fisher}}
Elastic weight consolidation \citep{kirkpatrick2017overcoming} was proposed to avoid catastrophic forgetting in the setup of lifelong learning, where several tasks should be learned sequentially. 
In addition to preserving the initial parameter vector $\initweights$, it consists in using the estimated Fisher information to define the distance between $\sourceweights$ and $\sourceinitweights$. 
More precisely, it relies on the diagonal of the Fisher information matrix, resulting in the following penalty:
\begin{equation}\label{eq:L2-SP-Fisher-full}
  \Omega(\weights) = \frac{\alpha}{2} \sum_{j\in\source} \hat{F}_{jj} 
  \left(\sourceweight_j-\sourceinitweight_j\right)^2 + \frac{\beta}{2} 
  \norm{\newweights}^2   
  \enspace,
\end{equation}
where $\hat{F}_{jj}$ is the estimate of the $j$th diagonal element of the Fisher information matrix.
It is computed as the average of the squared Fisher's score on the source problem, using the inputs of the source data:
\begin{equation*}
  \hat{F}_{jj} = \frac{1}{m} \sum_{i=1}^m \sum_{k=1}^K \funcnet(\xinputs;\initweights) \left( \frac{\partial}{\partial\weight_j}\log \funcnet(\xinputs;\initweights) \right)^2
  \hspace{-0.5ex},
\end{equation*}
where the outer average estimates the expectation with respect to inputs 
$\xinputs[]$ and the inner weighted sum is the estimate of the conditional expectation of outputs given input $\xinputs$, with outputs drawn from a categorical distribution of parameters $(\funcnet[1](\xinputs;\weights), \ldots, \funcnet(\xinputs;\weights), \ldots, \funcnet[K](\xinputs;\weights))$.

\paragraph{\boldmath$L^1$\textit{-SP}}
We also experiment the $L^1$ variant of $L^2$\textit{-SP}:
\begin{equation}\label{eq:L1-SP-full}
  \Omega(\weights) = \alpha\norm[1]{\sourceweights-\sourceinitweights} + \frac{\beta}{2} \norm{\newweights}^2
  \enspace.
\end{equation}
The usual $L^1$ penalty encourages sparsity; here, by using $\sourceinitweights$ as a reference in the penalty, $L^1$-SP encourages some components of the parameter vector to be frozen, equal to the pre-trained initial values. 
The penalty can thus be thought as intermediate between $L^2$\textit{-SP} \eqref{eq:L2-SP-full} and the strategies consisting in freezing a part of the initial network.
We explore below other ways of doing so.

\paragraph{Group-Lasso\textit{-SP} (\textit{GL-SP})}
Instead of freezing some individual parameters, we may encourage freezing some groups of parameters corresponding to channels of convolution kernels.
Formally, we endow the set of parameters with a group structure, defined by a fixed partition
of the index set
$\mathcal{I}=\{1,\ldots,p\}$, that is,
$
  \mathcal{I}=\bigcup_{g=0}^G\group,\, \text{with}\enspace 
  \group \cap \group[h]=\emptyset \enspace
  \text{for}\enspace g\neq h.
$
In our setup, $\group[0] = \bar{\source}$, and for $g>0$, $\group$ is the set of fan-in parameters of channel $g$.
Let 
$p_g$ denote the cardinality of group $g$, and 
$\weights_{\group}$
$\in\Rset^{p_g}$ 
be the vector $(\weight_j)_{j\in \group}$.
Then, the \textit{GL-SP} penalty is:
\begin{equation}\label{eq:group-lasso-sp}
  \Omega(\weights) = \alpha \sum_{g=1}^G s_g \norm{\weights_{\group} - \initweights_{\group}} + \frac{\beta}{2} \norm{\weights_{\bar{\source}}}^2
  \enspace,
\end{equation}
where $\initweights_{\group[0]} = \initweights_{\bar{\source}} \eqdef \boldsymbol{0}$, and, for $g>0$, $s_g$ is a predefined constant that may be used to balance the different cardinalities of groups.
In our experiments, we used $s_g=p_{g}^{1/2}$.

Our implementation of Group-Lasso\textit{-SP} can freeze feature extractors at any depth of the convolutional network, to preserve the pre-trained feature extractors as a whole instead of isolated pre-trained parameters.
The group $\group$ of size $p_{g}=h_{g}\times \mathrm{w}_{g} \times d_{g}$ gathers all the parameters of a convolution kernel of height $h_{g}$, width $\mathrm{w}_{g}$, and depth $d_{g}$.
This grouping is done at each layer of the network, for each output channel, so that the group index $g$ corresponds to two indexes in the network architecture: the layer index $l$ and the output channel index at layer $l$. 
If we have $c_{l}$ such channels at layer $l$, we have a total of $G = \sum_{l} c_l$ groups.

\paragraph{Group-Lasso-SP-Fisher (\textit{GL-SP-Fisher})}
Following the idea of $L^2$\textit{-SP-Fisher}, the Fisher version of \textit{GL-SP} is:

\begin{equation*}
  \Omega(\weights) = 
  \alpha \sum_{g=1}^G s_g  \Big( \sum_{j\in\group}\hat{F}_{jj} \left(\sourceweight_j-\sourceinitweight_j\right)^2\Big)^{1/2}\hspace{-1.25ex}+\hspace{-0.125ex}
  \frac{\beta}{2} \norm{\weights_{\group[0]}}^2
  \hspace{0.5ex}.
\end{equation*}

\section{Experiments}
\label{sec:experiments}

We evaluate the aforementioned parameter regularizers on several pairs of source and target tasks.
We use ResNet \citep{he2016deep} 
as our base network, since it has proven its wide applicability on transfer learning tasks.
Conventionally, if the target task is also a classification task, the training process starts by replacing the last layer with a new one, randomly generated, whose size depends on the number of classes in the target task.

\subsection{Source and Target Databases}

For comparing the effect of similarity between 
the source problem and the target problem on transfer learning, we chose two source databases:
ImageNet \citep{deng2009imagenet} for generic object recognition and Places 365 \citep{zhou2017places} for scene classification.
%
Likewise, we have three different databases related to three target problems:
Caltech 256 \citep{griffin2007caltech} contains different objects for generic object recognition;
MIT Indoors 67 \citep{quattoni2009recognizing} consists of 67 indoor scene categories;
Stanford Dogs 120 \citep{khosla2011novel} contains images of 120 breeds of dogs;
Each target database is split into training and testing sets following the suggestion of their creators
(see Table~\ref{table:db} for details).
In addition, we consider two configurations for Caltech 256: 30 or 60 examples randomly drawn from each category for training, and 20 remaining examples for test.

\begin{table*}
	\centering
	\caption{Characteristics of the target databases: name and type, numbers of training and test images per class, and number of classes.}
	\begin{tabular}{ c | c | c | c | c }
		Database & task category& \# training &  \# test & \# classes    \rule{0ex}{2.25ex} \\[.1ex] \hline \rule{0ex}{2.25ex}
		MIT Indoors 67    & scene classification        & ~\,80 &    ~~~ 20 & ~\,67 \\
		Stanford Dogs 120 & specific object recog. & 100 & $\sim$ 72   & 120   \\
		Caltech 256 -- 30 & generic object recog.  & ~\,30 & ~~~ 20  & 257   \\
		Caltech 256 -- 60 & generic object recog.  & ~\,60 & ~~~ 20    & 257   \\
	\end{tabular}
	\label{table:db}
\end{table*}

\subsection{Training Details}

\setlength\figH{5cm}
\setlength\figW{10cm}
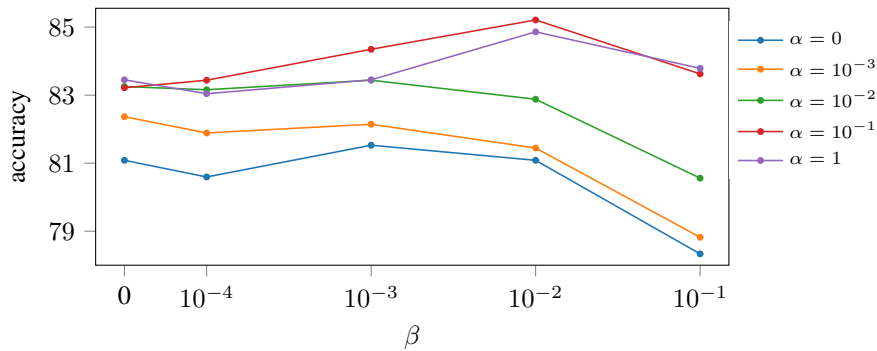
\begin{figure*}[t]
\centering
\begin{tikzpicture}

\definecolor{color1}{rgb}{1,0.498039215686275,0.0549019607843137}
\definecolor{color0}{rgb}{0.12156862745098,0.466666666666667,0.705882352941177}
\definecolor{color3}{rgb}{0.83921568627451,0.152941176470588,0.156862745098039}
\definecolor{color2}{rgb}{0.172549019607843,0.627450980392157,0.172549019607843}
\definecolor{color4}{rgb}{0.580392156862745,0.403921568627451,0.741176470588235}

\begin{axis}[
xlabel={$\beta$},
compat=newest,
ylabel={accuracy},
xmin=-0.175, xmax=3.675,
ymin=78, ymax=85.5536237359047,
width=\figW,
height=\figH,
xtick={0,0.5,1.5,2.5,3.5},
xticklabels={0,$10^{-4}$,$10^{-3}$,$10^{-2}$,$10^{-1}$},
ytick={79,81,83,85},
tick align=outside,
tick pos=left,
x grid style={lightgray!92.026143790849673!black},
y grid style={lightgray!92.026143790849673!black},
legend style={font=\fontsize{7}{5}\selectfont, at={(1.0,0.95)}, anchor=north west, draw=none},
legend cell align={left},
legend entries={{$\alpha=0$},{$\alpha=10^{-3}$},{$\alpha=10^{-2}$},{$\alpha=10^{-1}$},{$\alpha=1$}}
]
\addplot [semithick, color0, mark=*, mark size=1, mark options={solid}]
table {%
0 81.0838997364044
0.5 80.594402551651
1.5 81.5267980098724
2.5 81.0838997364044
3.5 78.3333003520966
};
\addplot [semithick, color1, mark=*, mark size=1, mark options={solid}]
table {%
0 82.3660016059875
0.5 81.8880975246429
1.5 82.1444988250732
2.5 81.4451992511749
3.5 78.82279753685
};
\addplot [semithick, color2, mark=*, mark size=1, mark options={solid}]
table {%
0 83.2517981529236
0.5 83.1584990024567
1.5 83.4381997585297
2.5 82.8787982463837
3.5 80.5594027042389
};
\addplot [semithick, color3, mark=*, mark size=1, mark options={solid}]
table {%
0 83.2167983055115
0.5 83.4381997585297
1.5 84.3473017215729
2.5 85.2097988128662
3.5 83.6247026920319
};
\addplot [semithick, color4, mark=*, mark size=1, mark options={solid}]
table {%
0 83.4499001502991
0.5 83.0420017242432
1.5 83.4499001502991
2.5 84.860098361969
3.5 83.7879002094269
};
\end{axis}

\end{tikzpicture}\\
\caption{Classification accuracy (in \%) on Stanford Dogs 120 for $L^2$\textit{-SP}, according to the two regularization hyperparameters $\alpha$ and $\beta$ respectively applied to the layers inherited from the source task and the last classification layer (see Equation \ref{eq:L2-SP-full}).}
\label{fig:wdrate}
\end{figure*}

Most images in those databases are color images.
If not, we create a three-channel image by duplicating the gray-scale data.
All images are pre-processed: we resize images to 256$\times$256 and subtract the mean activity computed over the training set from each channel, then we adopt random blur, random mirror and random crop to 224$\times$224 for data augmentation.
The network parameters are regularized as described in Section \ref{section:ta}.
Cross validation is used for searching the best regularization hyperparameters $\alpha$ and $\beta$:
$\alpha$ differs across experiments, and $\beta=0.01$ is consistently picked by cross-validation for regularizing the last layer.
Figure \ref{fig:wdrate} illustrates that the test accuracy varies smoothly according to the regularization strength, and that there is a sensible benefit in penalizing the last layer (that is, $\beta \geq 0$) for the best $\alpha$ values.
When applicable, the Fisher information matrix is estimated on the source database.
The two source databases (ImageNet or Places 365) yield different estimates.
Regarding testing, we use central crops as inputs to compute the classification accuracy.

Stochastic gradient descent with momentum 0.9 is used for optimization.
We run 9000 iterations and divide the learning rate by 10 after 6000 iterations.
The initial learning rates are 0.005, 0.01 or 0.02, depending on the tasks.
Batch size is 64.
Then, under the best configuration, we repeat five times the learning process to obtain an average classification accuracy and standard deviation.
All the experiments are performed with Tensorflow \citep{tensorflow2015-whitepaper}.

\subsection{Results}
\label{sec:experiments_results}

\subsubsection{Fine-Tuning from a Similar Source}

\begin{table*}
\caption{Average classification accuracies (in \%) of $L^2$, $L^2$\textit{-SP} and $L^2$\textit{-SP-Fisher} on 5 different runs.
	The source database is Places 365 for MIT Indoors 67 and ImageNet for Stanford Dogs 120 and Caltech 256.}
  \begin{center}
    \begin{tabular}{c|c|c|c|c}
  & MIT Indoors 67 & Stanford Dogs 120  & Caltech 256 -- 30 & Caltech 256 -- 60  \\[.1ex] \hline \rule{0ex}{2.25ex}
  $L^2$           & 79.6$\pm$0.5 & 81.4$\pm$0.2  & 81.5$\pm$0.2 & 85.3$\pm$0.2\\
  $L^2$\textit{-SP}        & \textbf{84.2$\pm$0.3} & \textbf{85.1$\pm$0.2} & \textbf{83.5$\pm$0.1} & \textbf{86.4$\pm$0.2} \\
  $L^2$\textit{-SP-Fisher} & 84.0$\pm$0.4 & \textbf{85.1$\pm$0.2}  & 83.3$\pm$0.1 & 86.0$\pm$0.1 \\
  \end{tabular}
  \end{center}
  \label{table:results}
\end{table*}

Table~\ref{table:results} displays the results of fine-tuning with $L^2$\textit{-SP} and $L^2$\textit{-SP-Fisher}, which are compared to the current baseline of fine-tuning with $L^2$.
We report the average accuracies and their standard deviations on 5 different runs.
Since we use the same data and the same starting point, runs differ only due to the randomness of stochastic gradient descent and to the weight initialization of the last layer.
We can observe that $L^2$\textit{-SP} and $L^2$\textit{-SP-Fisher} always improve over $L^2$, and that when less training data are available for the target problem, the improvement of $L^2$\textit{-SP} and $L^2$\textit{-SP-Fisher} compared to $L^2$ are more important.
Meanwhile, no large difference is observed between $L^2$\textit{-SP} and $L^2$\textit{-SP-Fisher}.

We can boost the performance and outperform the state of the art \citep{ge2017borrowing} in some cases by exploiting more training techniques and post-processing methods, which are described in the supplementary material.

\subsubsection{Behavior Across Penalties, Source and Target Databases}

\setlength\figH{4cm}
\setlength\figW{0.58\columnwidth}
\setlength\textSize{5.5pt}
\begin{figure}[t]
\centering
\begin{tikzpicture}

\begin{groupplot}[group style={group size=2 by 2,vertical sep=1.5cm},height=\figH,width=\figW]
\nextgroupplot[
xmin=-0.25, xmax=5.25,
ymin=0.74, ymax=0.86,
title = MIT Indoor 67,
xtick={0,1,2,3,4,5},
ytick={0.75,0.80,0.85},
xticklabels={$L^2$,$L^2$\textit{-SP},$L^2$\textit{-SP-F},$L^1$\textit{-SP},\textit{GL-SP},\textit{GL-SP-F}},
yticklabels={75,80,85},
tick align=outside,
xticklabel style = {rotate=20, font=\fontsize{\textSize}{2}\selectfont},
yticklabel style = {font=\fontsize{6}{2}\selectfont},
tick pos=left,
x grid style={lightgray!92.026143790849673!black},
y grid style={lightgray!92.026143790849673!black},
legend style={at={($(0,0)+(1cm,1cm)$)},legend columns=2,fill=none,draw=black,anchor=center,align=center},
]
\addplot [semithick, blue, mark=*, mark size=1, mark options={solid}, only marks, forget plot]
table {%
0 0.767164
0 0.778358
0 0.776119
0 0.764179
0 0.767164
1 0.780597
1 0.778358
1 0.774627
1 0.780597
1 0.773134
2 0.781343
2 0.782836
2 0.783582
2 0.782836
2 0.777612
3 0.778358
3 0.78209
3 0.773134
3 0.775373
3 0.782836
4 0.78806
4 0.783582
4 0.786567
4 0.784328
4 0.78209
5 0.78209
5 0.773881
5 0.78209
5 0.772388
5 0.776119
};
\addplot [semithick, red, mark=*, mark size=1, mark options={solid}, only marks, forget plot]
table {%
0 0.797761
0 0.802239
0 0.78806
0 0.797761
0 0.79403
1 0.846269
1 0.839552
1 0.841045
1 0.841791
1 0.842537
2 0.842537
2 0.834328
2 0.84403
2 0.838806
2 0.83806
3 0.837313
3 0.835075
3 0.83806
3 0.830597
3 0.833582
4 0.842537
4 0.852985
4 0.843284
4 0.849254
4 0.85
5 0.841791
5 0.843284
5 0.84403
5 0.838806
5 0.842537
};
\nextgroupplot[
xmin=-0.25, xmax=5.25,
title=Stanford Dogs 120,
xtick={0,1,2,3,4,5},
ytick={0.74,0.78,0.82,0.86},
xticklabels={$L^2$,$L^2$\textit{-SP},$L^2$\textit{-SP-F},$L^1$\textit{-SP},\textit{GL-SP},\textit{GL-SP-F}},
tick align=outside,
yticklabels={74,78,82,86},
xticklabel style = {rotate=20, font=\fontsize{\textSize}{2}\selectfont},
yticklabel style = {font=\fontsize{6}{2}\selectfont},
tick pos=left,
x grid style={lightgray!92.026143790849673!black},
y grid style={lightgray!92.026143790849673!black}
]
\addplot [semithick, blue, mark=*, mark size=1, mark options={solid}, only marks, forget plot]
table {%
0 0.812471
0 0.812238
0 0.814452
0 0.815152
0 0.815618
1 0.852214
1 0.852797
1 0.849184
1 0.848951
1 0.851632
2 0.850466
2 0.852681
2 0.852098
2 0.849184
2 0.848951
3 0.851981
3 0.853147
3 0.854429
3 0.852215
3 0.850932
4 0.847552
4 0.850932
4 0.847902
4 0.851399
4 0.848019
5 0.851049
5 0.850466
5 0.851166
5 0.850117
5 0.849767
};
\addplot [semithick, red, mark=*, mark size=1, mark options={solid}, only marks, forget plot]
table {%
0 0.748252
0 0.752098
0 0.748368
0 0.748019
0 0.749184
1 0.816434
1 0.818765
1 0.817832
1 0.816434
1 0.819697
2 0.817949
2 0.817133
2 0.815734
2 0.814219
2 0.816434
3 0.806294
3 0.806177
3 0.807925
3 0.807459
3 0.807692
4 0.80676
4 0.808858
4 0.808858
4 0.809208
4 0.806993
5 0.806061
5 0.808625
5 0.807226
5 0.805012
5 0.809557
};
\nextgroupplot[
xmin=-0.25, xmax=5.25,
width=\figW,
height=\figH,
title= Caltech 256 -- 30,
xtick={0,1,2,3,4,5},
ytick={0.78,0.81,0.84},
xticklabels={$L^2$,$L^2$\textit{-SP},$L^2$\textit{-SP-F},$L^1$\textit{-SP},\textit{GL-SP},\textit{GL-SP-F}},
tick align=outside,
yticklabels={78,81,84},
xticklabel style = {rotate=20, font=\fontsize{\textSize}{2}\selectfont},
yticklabel style = {font=\fontsize{6}{2}\selectfont, /pgf/number format/precision=2},
tick pos=left,
x grid style={lightgray!92.026143790849673!black},
y grid style={lightgray!92.026143790849673!black}
]
\addplot [semithick, blue, mark=*, mark size=1, mark options={solid}, only marks,forget plot]
table {%
0	0.817647
0	0.812257
0	0.812745
0	0.816732
0	0.814706
1	0.834706
1	0.835294
1	0.834118
1	0.83549
1	0.833333
2	0.832685
2	0.83249
2	0.833268
2	0.833268
2	0.835294
3	0.83035
3	0.832296
3	0.830156
3	0.831712
3	0.831176
4	0.832685
4	0.83249
4	0.833268
4	0.833268
4	0.835294
5	0.83035
5	0.830156
5	0.830739
5	0.831128
5	0.832296
};
\addplot [semithick, red, mark=*, mark size=1, mark options={solid}, only marks, forget plot]
table {%
0	0.78249
0	0.780739
0	0.782101
0	0.787549
0	0.783268
1	0.802335
1	0.800973
1	0.803502
1	0.801946
1	0.802335
2	0.788521
2	0.783852
2	0.7893
2	0.78677
2	0.792023
3	0.782879
3	0.781712
3	0.782685
3	0.782685
3	0.783658
4	0.790467
4	0.794553
4	0.793191
4	0.792218
4	0.792607
5	0.783074
5	0.781323
5	0.781517
5	0.781712
5	0.78249
};
\nextgroupplot[
xmin=-0.25, xmax=5.25,
ymin=0.815, ymax=0.87,
width=\figW,
height=\figH,
title= Caltech 256 -- 60,
xtick={0,1,2,3,4,5},
ytick={0.82,0.84,0.86},
yticklabel style={/pgf/number format/precision=3},
xticklabels={$L^2$,$L^2$\textit{-SP},$L^2$\textit{-SP-F},$L^1$\textit{-SP},\textit{GL-SP},\textit{GL-SP-F}},
tick align=outside,
yticklabels={82,84,86},
xticklabel style = {rotate=20, font=\fontsize{\textSize}{2}\selectfont},
yticklabel style = {font=\fontsize{6}{2}\selectfont},
tick pos=left,
x grid style={lightgray!92.026143790849673!black},
y grid style={lightgray!92.026143790849673!black},
legend to name=fan,
legend style={at={($(0,0)+(1cm,1cm)$)},legend columns=2,fill=none,draw=black,anchor=center,align=center,column sep=5pt},
]
\addplot [semithick, blue, mark=*, mark size=1, mark options={solid}, only marks]
table {%
	0	0.856031
	0	0.853333
	0	0.851556
	0	0.852549
	0	0.852549
	1	0.866275
	1	0.863922
	1	0.863333
	1	0.862353
	1	0.861373
	2	0.858366
	2	0.859533
	2	0.860506
	2	0.861479
	2	0.859922
	3	0.85856
	3	0.858171
	3	0.857198
	3	0.855253
	3	0.85856
	4	0.859339
	4	0.858949
	4	0.858755
	4	0.857198
	4	0.857977
	5	0.859339
	5	0.857782
	5	0.861673
	5	0.860506
	5	0.856615
};
\addplot [semithick, red, mark=*, mark size=1, mark options={solid}, only marks]
table {%
	0	0.831712
	0	0.836576
	0	0.831907
	0	0.833463
	0	0.835798
	1	0.838132
	1	0.835214
	1	0.838521
	1	0.838132
	1	0.839883
	2	0.835603
	2	0.838132
	2	0.83891
	2	0.834825
	2	0.83696
	3	0.825292
	3	0.827626
	3	0.829572
	3	0.827626
	3	0.827432
	4	0.82821
	4	0.827626
	4	0.831323
	4	0.828988
	4	0.82821
	5	0.826265
	5	0.826654
	5	0.829572
	5	0.828405
	5	0.82821
};
\end{groupplot}
\node[below] at (5.5,3.0) {\pgfplotslegendfromname{fan}};
\end{tikzpicture}\\
\caption{Classification accuracies (in \%) of the tested fine-tuning approaches on the four target databases, using ImageNet (dark blue dots) or Places 365 (light red dots) as source databases. MIT Indoor 67 is more similar to Places 365 than to ImageNet; Stanford Dogs 120 and Caltech 256 are more similar to ImageNet than to Places 365.}
\label{fig:results}
\end{figure}
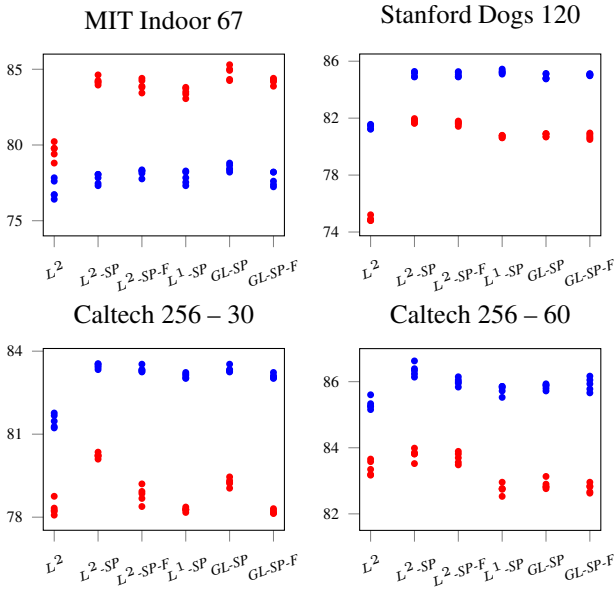

A comprehensive view of our experimental results is given in Figure \ref{fig:results}.
Each plot corresponds to one of the four target databases listed in Table~\ref{table:db}.
The light red points mark the accuracies of transfer learning when using Places 365 as the source database, whereas the dark blue points correspond to the results obtained with ImageNet.
As expected, the results of transfer learning are much better when source and target are alike:
the scene classification target task MIT Indoor 67 (top left) is better transferred from the scene classification source task  Places 365, whereas the object recognition target tasks benefit more from the object recognition source task ImageNet.
It is however interesting to note that the trends are similar for the two source databases:
all the fine-tuning strategies based on penalties using the starting point \textit{-SP} as a reference perform consistently better than standard fine-tuning ($L^2$).
There is thus a benefit in having an explicit bias towards the starting point, even when the target task is not too similar to the source task.

This benefit is comparable for $L^2$\textit{-SP} and $L^2$\textit{-SP-Fisher} penalties; the strategies based on $L^1$ and Group-Lasso penalties behave rather poorly in comparison.
They are even less accurate than the plain $L^2$ strategy on Caltech 256 -- 60 when the source problem is Places 365.
Stochastic gradient descent does not handle well these penalties whose gradient is discontinuous at the starting point where the optimization starts.
The stochastic forward-backward splitting algorithm of \citet{duchi2009efficient}, which is related to proximal methods, leads to substandard results, presumably due to the absence of a momentum term.
In the end, we used plain stochastic gradient descent on a smoothed version of the penalties eliminating the discontinuities of their gradients, but some instability remains.

Finally, the variants using the Fisher information matrix behave like the simpler variants using a Euclidean metric on parameters.
We believe that this is due to the fact that, contrary to lifelong learning, our objective does not favor solutions that retain accuracy on the source task.
The metric defined by the Fisher information matrix may thus be less relevant for our actual objective that only relates to the target task.
Table \ref{table:source-data} confirms that $L^2$\textit{-SP-Fisher} is indeed a better approach in the situation of lifelong learning, where accuracies on the source tasks matter. It reports the drop in performance when the fine-tuned models are applied on the source task, without any retraining, simply using the original classification layer instead of the classification layer learned for the target task.
The performance drop is smaller for $L^2$\textit{-SP-Fisher} than for $L^2$\textit{-SP}. 
In comparison, $L^2$ fine-tuning results in catastrophic forgetting: the performance on the source task is considerably affected by fine-tuning.

\begin{table}
	\caption{Classification accuracy drops (in \%) on the source tasks due to fine-tuning based on $L^2$, $L^2$\textit{-SP} and $L^2$\textit{-SP-Fisher} regularizers. 
		The source database is Places 365 for MIT Indoors 67 and ImageNet for Stanford Dogs 120 and Caltech 256.
		The classification accuracies of the pre-trained models are 54.7\% and 76.7\%  on Places 365 and ImageNet respectively.}
	\begin{center}
		\begin{tabular}{@{}c|c|c|c@{}}
			&  $L^2$ &  $L^2$\textit{-SP} &  $L^2$\textit{-SP-Fisher}  \\[.1ex] \hline 
			MIT Indoors 67 & -24.1  & -5.3 & -4.9\rule{0ex}{2.25ex}\\
			Stanford Dogs 120 & -14.1  & -4.7 & -4.2\\
			Caltech 256 -- 30 & -15.4  & -4.2 & -3.6\\
			Caltech 256 -- 60 & -16.9  & -3.6 & -3.2\\
		\end{tabular}
	\end{center}
	\label{table:source-data}
\end{table}

\subsubsection{Fine-Tuning {\em{vs.}} Freezing the Network}

Freezing the first layers of a network during transfer learning is another way to ensure a very strong inductive bias, letting less degrees of freedom to transfer learning.
Figure~\ref{fig:n_fixed_layers} shows that this strategy, which is costly to implement if one looks for the optimal number of layers to be frozen, can improve $L^2$ fine-tuning considerably, but that it is a rather inefficient strategy for $L^2$\textit{-SP} fine-tuning.
Among all possible choices, $L^2$ fine-tuning with partial freezing is dominated by the plain $L^2$\textit{-SP} fine-tuning.
Note that $L^2$\textit{-SP-Fisher} (not displayed) behaves similarly to $L^2$\textit{-SP}.

\setlength\figH{4.8cm}
\setlength\figW{8.25cm}
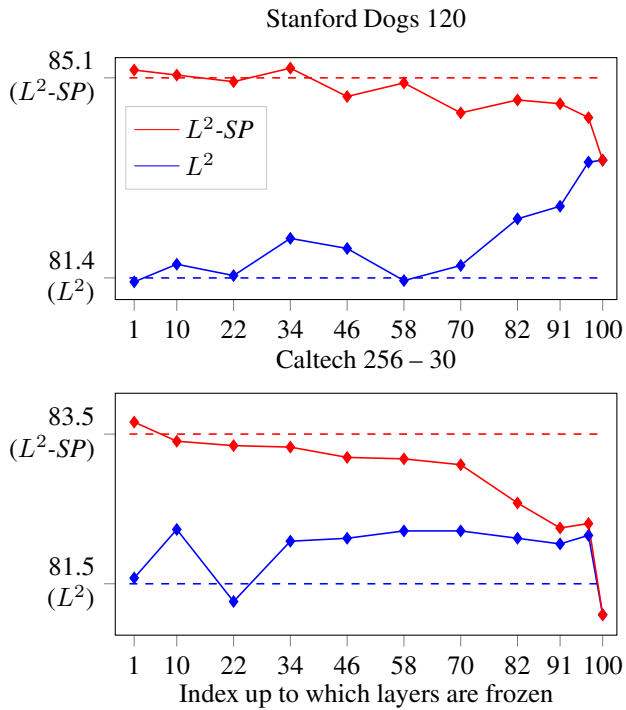
\begin{figure}[t]
\centering
\begin{tikzpicture}

\begin{groupplot}[group style={group size=1 by 2, vertical sep=1.25cm}]
\nextgroupplot[
xmin=-3, xmax=103,
ymin=0.81, ymax=0.854737375,
width=\figW,
height=\figH,
title=Stanford Dogs 120,
ytick={0.814,0.851},
yticklabels={81.4\\($L^2$),85.1\\($L^2$\textit{-SP})},
yticklabel style={align=right},
xtick={1,10,22,34,46,58,70,82,91,100},
xticklabels={1,10,22,34,46,58,70,82,91,100},
tick align=outside,
tick pos=left,
x grid style={lightgray!92.026143790849673!black},
y grid style={lightgray!92.026143790849673!black},
legend style={at={(0.02,0.80)}, anchor=north west, draw=white!80.0!black},
legend cell align={left},
legend entries={{$L^2$\textit{-SP}},{$L^2$}}
]
\addlegendimage{no markers, red}
\addlegendimage{no markers, blue}
\addplot [semithick, blue, dashed, forget plot]
table {%
0 0.814
99 0.814
};
\addplot [semithick, red, dashed, forget plot]
table {%
0 0.851
100 0.851
};
\addplot [semithick, blue, mark=diamond*, mark size=2, mark options={solid}]
table {%
1 0.813268
10 0.816531
22 0.814434
34 0.82131
46 0.819445
58 0.813501
70 0.816298
82 0.824923
91 0.827254
97 0.835413
100 0.835781
};
\addplot [semithick, red, mark=diamond*, mark size=2, mark options={solid}]
table {%
1 0.8524475
10 0.8515155
22 0.8502915
34 0.8527975
46 0.847552
58 0.850058
70 0.8445225
82 0.8469115
91 0.846212
97 0.843648
100 0.835781
};
\nextgroupplot[
title= Caltech 256 -- 30,
xlabel={Index up to which layers are frozen},
xmin=-3, xmax=103,
ymin=0.808, ymax=0.84,
width=\figW,
height=\figH,
ytick={0.8148,0.8346},
yticklabels={81.5\\($L^2$),83.5\\($L^2$\textit{-SP})},
yticklabel style={align=right},
xtick={1,10,22,34,46,58,70,82,91,100},
xticklabels={1,10,22,34,46,58,70,82,91,100},
tick align=outside,
tick pos=left,
x grid style={lightgray!92.026143790849673!black},
y grid style={lightgray!92.026143790849673!black},
]
\addplot [semithick, blue, dashed, forget plot]
table {%
0 0.8148
99 0.8148
};
\addplot [semithick, red, dashed, forget plot]
table {%
0 0.8346
99 0.8346
};
\addplot [semithick, blue, mark=diamond*, mark size=2, mark options={solid}]
table {%
1	0.815564
10	0.821984
22	0.812451
34	0.820428
46	0.820817
58	0.82179
70	0.82179
82	0.820817
91	0.82007
97	0.821206
100 0.8107
};
\addplot [semithick, red, mark=diamond*, mark size=2, mark options={solid}]
table {%
1	0.836187
10	0.833658
22	0.833074
34	0.832879
46	0.831517
58	0.831323
70	0.830545
82	0.825486
91	0.822179
97	0.822763
100 0.8107
};
\end{groupplot}

\end{tikzpicture}\\
\caption{Classification accuracies (in \%) of fine-tuning with $L^2$ and $L^2$\textit{-SP} on Stanford Dogs 120 (top) and Caltech 256--30 (bottom) when freezing the first layers of ResNet-101.
The dashed lines represent the accuracies reported in Table~\ref{table:results}, where no layers are frozen.
ResNet-101 begins with one convolutional layer, then stacks 3-layer blocks. 
The three layers in one block are either frozen or trained altogether.}
\label{fig:n_fixed_layers}
\end{figure}

\subsection{Analysis and Discussion}

Among all \textit{-SP} methods, 
$L^2$\textit{-SP} and $L^2$\textit{-SP-Fisher} always reach a better accuracy on the target task.
We expected $L^2$\textit{-SP-Fisher} to outperform $L^2$\textit{-SP}, since Fisher information helps in lifelong learning, but there is no significant difference between the two options.
Since $L^2$\textit{-SP} is simpler than $L^2$\textit{-SP-Fisher}, we recommend the former, and we focus on the analysis of $L^2$\textit{-SP}, although most of the analysis and the discussion would also apply to $L^2$\textit{-SP-Fisher}.

\subsubsection{Computational Efficiency}

The \textit{-SP} penalties introduce no extra parameters, and they only increase slightly the computational burden.
$L^2$\textit{-SP} increases the number of floating point operations required for a learning step of ResNet-101 by less than 1\%.
Hence, at a negligible computational cost, we can obtain significant improvements in classification accuracy, and no additional cost is experienced at test time.

\subsubsection{Theoretical Insights}

\setlength\figH{6cm}
\setlength\figW{15cm}
\begin{figure*}[t]
\centering
\input{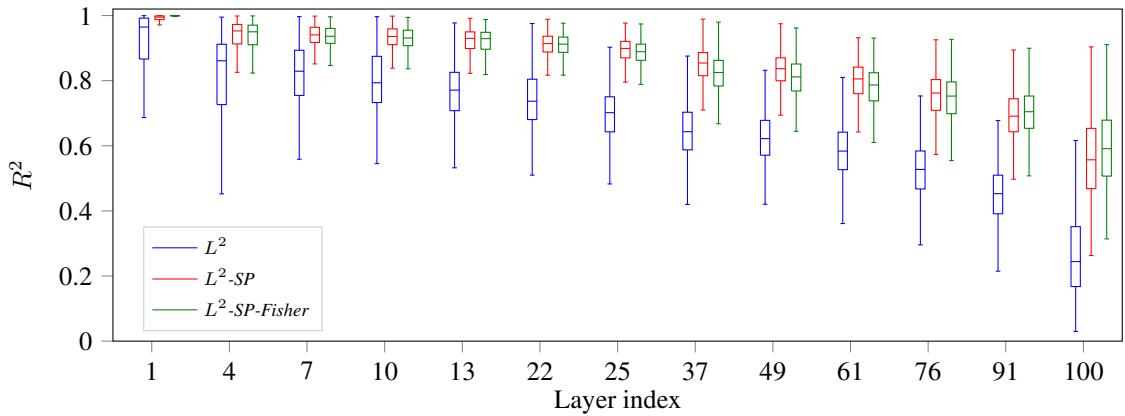}\\
\caption{$R^2$ coefficients of determination with $L^2$ and $L^2$\textit{-SP} regularizations for Stanford Dogs 120. Each boxplot summarizes the distribution of the $R^2$ coefficients of the activations after fine-tuning with respect to the activations of the pre-trained network, for all the units in one layer. ResNet-101 begins with one convolutional layer, then stacks 3-layer blocks. For legibility, we only display here the  $R^2$ at the first layer and at the outputs of some 3-layer blocks.}
\label{fig:dogs_r2}
\end{figure*}

Analytical results are very difficult to obtain in the deep learning framework.
Under some (highly) simplifying assumptions, we show in supplementary material that the optimum of the regularized objective function with $L^2$\textit{-SP} is a compromise between the optimum of the unregularized objective function and the pre-trained parameter vector, precisely an affine combination along the directions of eigenvectors of the Hessian matrix of the unregularized objective function.
This contrasts with $L^2$ that leads to a compromise between the optimum of the unregularized objective function and the origin.
Clearly, searching for a solution in the vicinity of the pre-trained parameters is intuitively much more appealing, since it is the actual motivation for using the pre-trained parameters as the starting point of the fine-tuning process.

Using $L^2$\textit{-SP} instead of $L^2$ can also be motivated by an analogy with shrinkage estimation \citep[see e.g.][chapter 5]{lehmann1998theory}. 
Although it is known that shrinking toward any reference is better than raw fitting, it is also known that shrinking towards a value that is close to the ``true parameters'' is more effective.
The notion of ``true parameters'' is not readily applicable to deep networks, but the connection with Stein shrinking effect may be inspiring by surveying the literature considering shrinkage towards other references, such as linear subspaces.
In particular, it is likely that manifolds of parameters defined from the pre-trained network would provide a more relevant reference than the single parameter value provided by the pre-trained network.

\subsubsection{Layer-Wise Analysis}

We complement our experimental results by an analysis relying on the activations of the hidden units of the network, to provide another view on the differences between $L^2$ and $L^2$\textit{-SP} fine-tuning.
Activation similarities are easier to interpret than parameter similarities, and they provide a view of the network that is closer to the functional perspective we are actually pursuing.
Matching individual activations makes sense, provided that the networks slightly differ before and after tuning so that few roles are switched between units or feature maps.

The dependency between the pre-trained and the fine-tuned activations throughout the network is displayed in Figure~\ref{fig:dogs_r2}, with boxplots of the $R^2$ coefficients, gathered layer-wise, of the fine-tuned activations with respect to the original activations.
This figure shows that, indeed, the roles of units or feature maps have not changed much after $L^2$\textit{-SP} and $L^2$\textit{-SP-Fisher} fine-tuning.
The $R^2$ coefficients are very close to 1 on the first layers, and smoothly decrease throughout the network, staying quite high, around 0.6, for $L^2$\textit{-SP} and $L^2$\textit{-SP-Fisher} at the greatest depth. 
In contrast, for $L^2$ regularization, some important changes are already visible in the first layers, and the $R^2$ coefficients eventually reach quite low values  at the greatest depth.
This illustrates in details how the roles of the network units is remarkably retained with $L^2$\textit{-SP} and $L^2$\textit{-SP-Fisher} fine-tuning, not only for the first layers of the networks, but also for the last high-level representations before classification.

\section{Conclusion}

We described and tested simple regularization techniques for inductive transfer learning. They all encode an explicit bias towards the solution learned on the source task, resulting in a compromise with the pre-trained parameter that is coherent with the original motivation for fine-tuning.
All the regularizers evaluated here have been already used for other purposes or in other contexts, but we demonstrated their relevance for inductive transfer learning with deep convolutional networks.

We show that a simple $L^2$ penalty using the starting point as a reference, $L^2$\textit{-SP}, is useful, even if early stopping is used.
This penalty is much more effective than the standard $L^2$ penalty that is commonly used in fine-tuning.
It is also more effective and simpler to implement than the strategy consisting in freezing the first layers of a network. 
We provide theoretical hints and strong experimental evidence showing that $L^2$\textit{-SP} retains the memory of the features learned on the source database.
We thus believe that this simple $L^2$\textit{-SP} scheme should be considered as the standard baseline in inductive transfer learning, and that future improvements of transfer learning should rely on this baseline.

Besides, we tested the effect of more elaborate penalties, based on $L^1$ or Group-$L^1$ norms, or based on Fisher information.
None of the $L^1$ or Group-$L^1$ options seem to be valuable in the context of inductive transfer learning that we considered here, and using the Fisher information with $L^2$\textit{-SP} does not improve accuracy on the target task.
Different approaches, which implement an implicit bias at the functional level, alike \citet{li2017learning}, remain to be tested: being based on a different principle, their value should be assessed in the framework of inductive transfer learning.

\section*{Acknowledgments}
This work was carried out with the supports of the China Scholarship Council and of a PEPS grant through the DESSTOPT project jointly managed by the National Institute of Mathematical Sciences and their Interactions (INSMI) and the Institute of Information Science and their Interactions (INS2I) of the CNRS, France. We acknowledge the support of NVIDIA Corporation with the donation of GPUs used for this research.

\bibliography{using}
\bibliographystyle{icml2018}

\clearpage
\appendix
\section{Effect of $L^2$\textit{-SP} Regularization on Optimization}
\label{appendix:l2-sp}

The effect of $L^2$ regularization can be analyzed by doing a quadratic approximation of the objective function around the optimum \citep[see, e.g.][Section 7.1.1]{Goodfellow-et-al-2017-Book}.
This analysis shows that $L^2$ regularization rescales the parameters along the directions defined by the eigenvectors of the Hessian matrix.
This scaling is equal to $\frac{\lambda_i}{\lambda_i+\alpha}$ for the $i$-th eigenvector of eigenvalue $\lambda_i$. 
A similar analysis can be used for the $L^2$\textit{-SP} regularization.

We recall that $J(\weights)$ is the unregularized objective function, and $\tilde{J}(\weights) = J(\weights)+\alpha \norm{\weights - \initweights}^2$ is the regularized objective function.
Let $\weights^* = \mathrm{argmin}_{\weights} J(\weights)$ and $\tilde{\weights} = \mathrm{argmin}_{\weights} \tilde{J}$ be their respective minima.
The quadratic approximation of $J(\weights^*)$ gives

\begin{equation}
  \label{eq:l2speffect}
  \mathbf{H}(\tilde{\weights} - \weights^*) + \alpha (\tilde{\weights} - \initweights) = 0
  \enspace,
\end{equation}
where $\mathbf{H}$ is the Hessian matrix of $J$ w.r.t. $\weights$, evaluated at $\weights^*$.
Since $\mathbf{H}$ is positive semidefinite, it can be decomposed as $\mathbf{H} = \mathbf{Q}\mathbf{\Lambda}\mathbf{Q}^T$. 
Applying the decomposition to Equation (\ref{eq:l2speffect}), we obtain the following relationship between $\tilde{\weights}$ and $\weights^*$:

\begin{equation}
\mathbf{Q}^T\tilde{\weights} = (\mathbf{\Lambda}+\alpha \mathbf{I})^{-1}\mathbf{\Lambda}\mathbf{Q}^T \weights^* + \alpha (\mathbf{\Lambda}+\alpha \mathbf{I})^{-1}\mathbf{Q}^T\initweights
\enspace.
\end{equation}

We can see that with $L^2$\textit{-SP} regularization, in the direction defined by the $i$-th eigenvector of $\mathbf{H}$, $\tilde{\weights}$  is a convex combination of $\weights^*$ and $\initweights$ in that direction since $\frac{\lambda_i}{\lambda_i+\alpha}$ and $\frac{\alpha}{\lambda_i+\alpha}$ sum to $1$.

\section{Matching the State of the Art in Image Classification}
\label{section:ap1}
The main objective of this paper is to demonstrate that \textit{-SP} regularization in general, and $L^2$\textit{-SP} in particular, provides a baseline for transfer learning that is significantly superior to the standard fine-tuning technique. 
We do not aim at reaching the state of the art solely with this simple technique.
However, as shown here, with some training tricks and post-processing methods, which have been proposed elsewhere but were not used in the paper, we can reach or even exceed the state of the art performances, simply by changing the regularizer to $L^2$\textit{-SP}.

\paragraph{Aspect Ratio.}
During training, respecting or ignoring the aspect ratio of images will give different results, and usually it would be better to keep the original aspect ratio.
In the paper, the classification experiments are all under the pre-processing of resizing all images to 256$\times$256, i.e. ignoring the aspect ratio.
Here we perform an ablation study to analyze the difference between keeping and ignoring the ratio.
For simplicity, we use the same hyperparameters as before except that the aspect ratio is kept and images are resized with the shorter edge being 256.

\paragraph{Post-Processing for Image Classification.}
A common post-processing method for image classification is 10-crop testing (averaging the predictions of 10 cropped patches, the four corner patches and the center patch as well as their horizontal reflections).

We apply the aspect ratio and 10-crop testing techniques to improve our results, but we believe the performance can be improved but using additional tricks, such as random rotation or scaling during training, more crops, multi scales for test, etc.
Table \ref{table:app-results} shows our results.
Caltech 256 - 30 outperforms the state of the art; our results in MIT Indoors 67 and Stanford Dogs 120 are very close to the state of the art, noting that the best performing approach \citep{ge2017borrowing} used many training examples from source domain to improve performance.
On our side, we did not use any other examples and simply changed the regularization approach from $L^2$ to $L^2$\textit{-SP}.

\begin{table*}[t]
	\centering
	\caption{Average classification accuracies (in \%) for $L^2$ and $L^2$\textit{-SP} using the training tricks presented in Section \ref{section:ap1}.
		The source database is Places 365 for MIT Indoors 67 and ImageNet for Caltech 256, Stanford Dogs and Foods.
		References for the state of the art are taken from \citet{ge2017borrowing}, except for Foods-101 where it is taken from \citet{martinel2016wide}.}
	\label{table:app-results}
	\begin{tabular}{c|c|c|c|c|c}
		
		& Caltech 256 - 30 & Caltech 256 - 60 & MIT Indoors 67 & Stanford Dogs 120 & Foods 101\\ \hline \rule{0ex}{2.25ex}
		$L^2$          & 82.7$\pm$0.2 & 86.5$\pm$0.4 & 80.7$\pm$0.9 & 83.1$\pm$0.2 & 86.7$\pm$0.2 \\
		$L^2$\textit{-SP} & 84.9$\pm$0.1 & 87.9$\pm$0.2 &  85.2$\pm$0.3 & 89.8$\pm$0.2 & 87.1$\pm$0.1 \\ \hline
		Reference & 83.8$\pm$0.5 & 89.1$\pm$0.2 & 85.8 & 90.3 & 90.3 \\
	\end{tabular}
\end{table*}

We add Foods 101 \citep{bossard14} to supplement our experiments.
Foods 101 is a database that collects photos of 101 food categories and is a much larger database than the three we already presented, yet rough in terms of image quality and class labels in the training set.

\section{Application of $L^2$\textit{-SP} to Semantic Image Segmentation}

The paper compares different regularization approaches for transfer in image classification.
In this section, we examine the versatility of $L^2$\textit{-SP} by applying it to image segmentation.
Although the image segmentation target task, which aims at labeling each pixel of an image with the category of the object it belongs to, differs from the image classification source task, it still benefits from fine-tuning.

We evaluate the effect of fine-tuning with $L^2$\textit{-SP} on Cityscapes \cite{Cordts2016Cityscapes}, a dataset with an evaluation benchmark for pixel-wise segmentation of real-world urban street scenes.
It consists of 5000 images with high quality pixel-wise labeling, which are split into a training set (2975 images), a validation set (500 images) and a test set (1525 images), all with resolution 2048$\times$1024 pixels.
ImageNet \cite{deng2009imagenet} is used as source.

As for the networks, we consider two architectures of convolutional networks: the standard ResNet \citep{he2016deep}, which can be used for image segmentation by removing the global pooling layer, and
DeepLab-V2 \citep{chen2017deeplab}, which stayed top-ranked for some time on the Cityscapes benchmark and is one of the most favored structures.
We reproduce them with $L^2$ and $L^2$\textit{-SP} on Cityscapes under the same setting.

Most of the training tricks used for classification apply to segmentation, and we precise here the difference.
Images are randomly cropped to 800$\times$800, 2 examples are used in a batch, and batch normalization layers are frozen to keep pre-trained statistics.
We use the polynomial learning rate policy as in \citet{chen2017deeplab} and the base learning rate is set to 0.0005.
For testing, we use the whole image.


\begin{table}[t]
	\caption{
		Mean IoU scores on Cityscapes validation set. Fine-tuning with $L^2$, \citet{chen2017deeplab} obtained 66.6 and 70.4 for ResNet-101 and DeepLab respectively.
	}
	\begin{center}
		\begin{tabular}{l | c | l }
			Method & $L^2$ & $L^2$\textit{-SP} \\[.1ex] \hline
			ResNet-101 & 68.1 & \textbf{68.7} \\
			DeepLab & 72.0 & \textbf{73.2} \\
		\end{tabular}
	\end{center}
	\label{table:results_seg}
\end{table}

Table~\ref{table:results_seg} reports the results on Cityscapes validation set.
We reproduce the experiments of ResNet and DeepLab that use the standard $L^2$ fine-tuning, and compare with $L^2$\textit{-SP} fine-tuning, all other setup parameters being unchanged.
We readily observe that fine-tuning with $L^2$\textit{-SP} in place of $L^2$ consistently improves the performance in mean IoU score, for both networks.


\end{document}